\documentclass[runningheads,oribibl]{llncs}
\usepackage[numbers,sort&compress]{natbib}
\usepackage[T1]{fontenc}
\usepackage{graphicx}%
\usepackage{multirow}%
\usepackage{amsmath,amssymb,amsfonts}%
\usepackage{mathrsfs}%
\usepackage[title]{appendix}%
\usepackage{xcolor}%
\usepackage{textcomp}%
\usepackage{manyfoot}%
\usepackage{booktabs}%
\usepackage[ruled,noend,nosemicolon,linesnumbered]{algorithm2e}
\usepackage{setspace}
\SetKwComment{Comment}{$\triangleright$\ }{}
\SetCommentSty{small}
\newlength{\commentWidth}
\newcommand{\FWCommentR}[1]{\Comment*[r]{\makebox[\commentWidth]{#1\hfill}}}
\newcommand{\FWCommentF}[1]{\Comment*[f]{\makebox[\commentWidth]{#1\hfill}}}
\usepackage{listings}%
\usepackage{pdflscape}

\usepackage{hyperref}
\usepackage{url}
\usepackage{cleveref}
\usepackage{siunitx}
\usepackage{subcaption}
\usepackage[skins,theorems]{tcolorbox}
\usepackage{orcidlink}
\usepackage{enumitem}

\spnewtheorem*{prooftothm2}{Proof to Theorem 2}{\itshape}{\rmfamily}

\newcommand{\fancynamehighlighted}{\textbf{F}eed-\textbf{F}orward with delayed \textbf{F}eedback}
\newcommand{\fancyabrv}{F\textsuperscript{3}}
\newcommand{\fancyabrvnosuperscript}{F3}

\begin{document}
\title{Feed-Forward Optimization With Delayed Feedback for Neural Network Training}
\titlerunning{Feed-Forward Optimization With Delayed Feedback}
%
\author{Katharina Flügel\inst{1,2}\and
Daniel Coquelin\inst{1,2}\and
Marie Weiel\inst{1,2}\and
Charlotte Debus\inst{1}\and
Achim Streit\inst{1}\and
Markus Götz\inst{1,2}}
\authorrunning{K. Flügel et al.}
%
\institute{Scientific Computing Center (SCC), Karlsruhe Institute of Technology (KIT), Karlsruhe, Germany,
\email{firstname.lastname@kit.edu} \and
Helmholtz AI, Germany}
\maketitle              
\begin{abstract}
Backpropagation has long been criticized for being biologically implausible due to its reliance on concepts that are not viable in natural learning processes. 
Two core issues are the weight transport and update locking problems caused by the forward-backward dependencies, which limit biological plausibility, computational efficiency, and parallelization. 
Although several alternatives have been proposed to increase biological plausibility, they often come at the cost of reduced predictive performance.
%
This paper proposes an alternative approach to training feed-forward neural networks addressing these issues by using approximate gradient information. 
We introduce \fancynamehighlighted{} (\fancyabrv{}), which approximates gradients using fixed random feedback paths and delayed error information from the previous epoch 
to balance biological plausibility with predictive performance.
%
We evaluate \fancyabrv{} across multiple tasks and architectures, including both fully-connected and Transformer networks. 
Our results demonstrate that, compared to similarly plausible approaches, \fancyabrv{} significantly improves predictive performance, narrowing the gap to backpropagation by up to 56\% for classification and 96\% for regression. 
This work is a step towards more biologically plausible learning algorithms while opening up new avenues for energy-efficient and parallelizable neural network training.
\let\thefootnote\relax\footnotetext{
The Version of Record of this contribution is published in the proceedings to the 31st International Conference on Neural Information Processing (ICONIP 2024), Lecture Notes in Computer Science, vol 15289, and is available online at 
\href{https://doi.org/10.1007/978-981-96-6585-3_6}{https://doi.org/10.1007/978-981-96-6585-3\_6}
}

\end{abstract}
\section{Introduction} 
\label{sec:introduction}
Today, nearly all artificial neural networks are trained with gradient-descent-based optimization methods~\citep{goodfellow_deep_2016} using backpropagation (BP)~\citep{rumelhart_learning_1986} to compute the gradients efficiently.
However, backpropagation relies on multiple biologically implausible factors, making it highly unlikely that the human brain learns in a similar fashion~\citep{bengio_towards_2016}.
While biological plausibility is not a necessity for training artificial neural networks, there has been great interest in methods that combine biological plausibility with effective training.
Two of the most substantial obstacles for these methods are the weight transport and the update locking problems~\citep{ororbia_biologically_2019,frenkel_learning_2021}.
%
\emph{Weight Transport}~\citep{grossberg_competitive_1987}:
Backpropagation reuses the forward weights symmetrically in the backward pass to propagate the gradients.
However, synapses are unidirectional, and synchronizing separate forward and backward pathways precludes biological plausibility~\citep{zeki_massively_2015,tang_bridging_2019}.
Additionally, the weights' non-locality constrains the memory access pattern, which can severely impact the computational efficiency~\citep{crafton_local_2019}.
\emph{Update Locking}~\citep{czarnecki_understanding_2017,jaderberg_decoupled_2017}:
Before computing the gradients in a backward pass, backpropagation requires a full forward pass to compute the network activations and the loss.
A layer thus needs to wait for all downstream layers before completing its own backward pass.
During this time, activations from the forward pass must be buffered or recomputed~\citep{mostafa_deep_2018,frenkel_learning_2021}.
This delay is not biologically plausible, and the forward-backward dependency restricts parallelization, for example, when pipelining over the network layers~\citep{huangGPipeEfficientTraining2019,narayananPipeDreamGeneralizedPipeline2019}.
Several alternatives have been proposed to solve these issues and increase biological plausibility, but they often come at the cost of reduced predictive performance.

\begin{figure*}[t]
    \vskip 0.2in
    \begin{center}
    \centerline{\includegraphics[width=\textwidth]{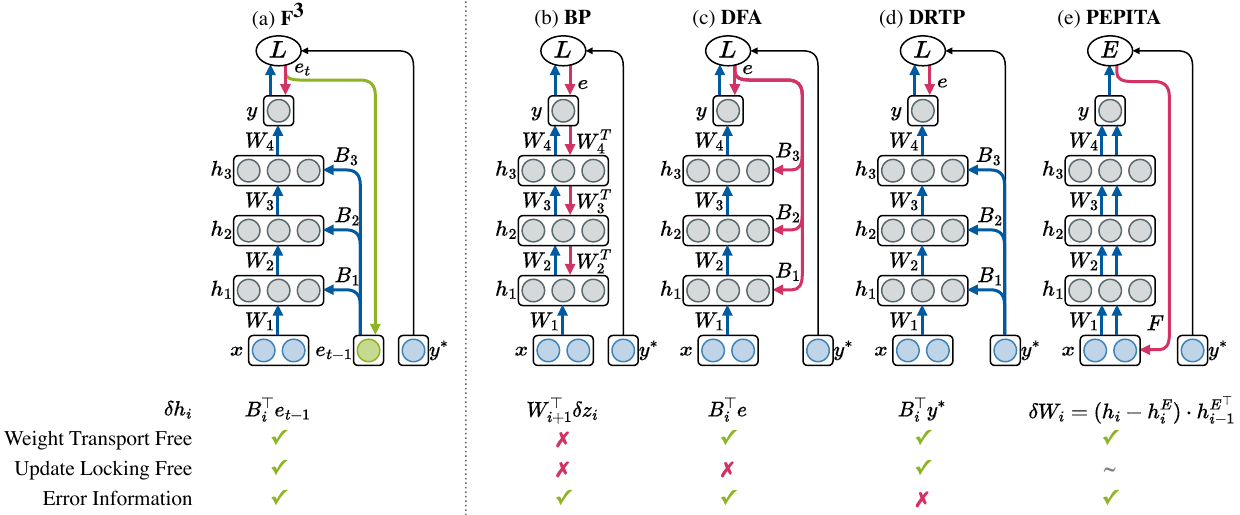}}
    \caption{
        \textbf{\fancyabrv{} (a)} solves both the weight transport and the update locking problems. In contrast to prior approaches, it uses delayed error information in the updates. The current error signal $e^t$ in epoch $t$ is stored (green) and used in the forward pass (blue) of the next epoch $t+1$, eliminating the backward pass (red) for all hidden layers. 
        \textbf{Previous approaches (b) to (e):} Backpropagation (BP) (b) is not biologically plausible due to the weight transport and update locking problems. DFA~\citep{nokland_direct_2016} (c) solves the weight transport problem by replacing the backward paths with direct random feedback paths $B_i$ but is still update-locked as it depends on the error $e$. DRTP~\citep{frenkel_learning_2021} (d) releases update locking by using the target $y^*$ instead of the error $e$, but this comes at the cost of reduced accuracy.
        PEPITA~\citep{dellaferrera_error-driven_2022} (e) improves the accuracy by using two forward passes per sample but is only partially update-unlocked.}
    \label{fig:ffo_concept_comparison}
    \end{center}
    \vspace{-0.75cm}
\end{figure*}

This paper presents \emph{\fancynamehighlighted{} (\fancyabrv)}, our novel, biologically inspired, backpropagation-free training algorithm for deep neural networks.
\fancyabrv{} yields superior predictive performance on various tasks compared to prior forward-only approaches by using delayed error information from previous epochs as feedback signals.
Using fixed feedback weights, \fancyabrv{} passes these feedback signals directly to each layer without layer-wise propagation.
Together, this resolves the dependence on downstream layers, enabling the update of network parameters during the forward pass.
As a result, \fancyabrv{} eliminates the need for the computationally expensive backward pass.
Besides increasing biological plausibility, this reduces the computational cost and consumed energy~\citep{crafton_local_2019,ororbia_biologically_2019} and opens up new avenues for parallelization by reducing the required communication between layers.
\Cref{fig:ffo_concept_comparison} illustrates \fancyabrv{} compared to backpropagation and prior biologically inspired approaches.
Our main contributions are as follows:
\begin{itemize}[itemsep=2pt, topsep=2pt]
    \item We introduce \fancyabrv{}, a bio-inspired, backpropagation-free training algorithm.
    \item We prove that \fancyabrv{}'s updates are in descending directions, minimizing the loss.
    \item We analyze the computational and memory complexity compared to BP.
    \item We evaluate \fancyabrv{} against other bio-plausible algorithms on various datasets and neural networks, including a proof-of-concept transformer network.
\end{itemize}
\section{Related Work} 
\label{sec:related_work}
Several approaches have been suggested to address both weight transport and update locking.
To solve the weight transport problem, \emph{target propagation}~\citep{lee_difference_2015} replaces the gradients with layer-wise target values computed by auto-encoders and a linear correction.
\citet{ororbia_biologically_2019} take a similar approach but employ the current layer's pre-activation and the next layer's post-activation instead of generating the targets with auto-encoders.
\emph{Feedback alignment (FA)}~\citep{lillicrap_random_2016} replaces the forward weights with fixed random feedback weights in the backward pass, demonstrating that symmetric feedback weights are not required for effective training.
However, the training signals remain non-local and are propagated backward through the network.
To alleviate this, \emph{direct feedback alignment (DFA)}~\citep{nokland_direct_2016} passes the error directly to each hidden layer, see \Cref{fig:ffo_concept_comparison} (c).

To solve update locking, \citet{mostafa_deep_2018} use auxiliary classifiers to generate \emph{local errors} from each layer's output.
Improving upon this, \citet{nokland_training_2019} demonstrate that combining local classifiers with a local similarity matching loss can close the gap to backpropagation.
\emph{Decoupled greedy learning (DGL)}~\citep{belilovsky_decoupled_2020} follows a similar approach that is based on greedy objectives, which even scales to large datasets such as ImageNet~\citep{deng_imagenet_2009}.
Another approach to address update locking are \emph{synthetic gradients}~\citep{jaderberg_decoupled_2017,czarnecki_understanding_2017}, which model a subgraph of the network and predict its future output based on only local information.
Replacing backpropagation with these synthetic gradients decouples the layers, resulting in \emph{decoupled neural interfaces (DNIs)}.
The same approach can also be used to predict synthetic input signals to solve the \emph{forward locking problem}~\citep{jaderberg_decoupled_2017}.
%
\emph{Direct random target projection (DRTP)}~\citep{frenkel_learning_2021}, illustrated in \Cref{fig:ffo_concept_comparison} (d), builds upon DFA to solve weight transport and update locking simultaneously.
By substituting the error with the targets, DFA's direct backward paths can be replaced with direct forward paths to the hidden layers.

The \emph{forward gradient}~\citep{baydin_gradients_2022} is another approach to gradient approximation using forward mode automatic differentiation in a single random direction.
The \emph{Forward-Forward (FF)} algorithm~\citep{hinton_forward_2022} replaces the backward pass with another forward pass on so-called negative data, training each layer to distinguish the positive from negative data.
\emph{PEPITA}~\citep{dellaferrera_error-driven_2022} takes a similar approach by executing two forward passes on different input data, where the input to the second forward pass is modulated based on the error of the first one.
\emph{Multilayer SoftHebb}~\citep{journe_hebbian_2022} combines Hebbian learning with a softmax-based plasticity rule and a corresponding soft anti-Hebbian plasticity to enable training deep networks of a specific architecture.
%
\emph{Predictive coding}~\citep{fristonTheoryCorticalResponses2005,raoPredictiveCodingVisual1999} is another approach to biological plausibility based on the predictive processing model.
It compares predictions of the expected input signal to a layer to the actual observations and aims to minimize this prediction error.
Predictive coding has been successfully applied to artificial neural networks, for example, in \citep{whittingtonApproximationErrorBackpropagation2017,millidgePredictiveCodingApproximates2022} and combined with Forward-Forward~\citep{ororbiaPredictiveForwardForwardAlgorithm2023a}.


\section{Feed-Forward-Only Training} 
\label{sec:neural_inspired_feed_forward_only_training_of_neural_networks}

\subsection{Training Neural Networks with Backpropagation}
\label{sub:backprop}
Let us consider a multi-layer neural network like the one illustrated in \Cref{fig:ffo_concept_comparison} consisting of $K$ fully-connected layers.
In the forward pass, each layer $i$ computes an activation $h_{i}=f(z_i)=f(W_{i}h_{i-1})$ using weights $W_i$ and a non-linear activation function $f$.
The loss function $L(y, y^*)$ measures how close the network's output $y=h_K$ for an input $x=h_0$ is to the corresponding target $y^*$.
Training a neural network aims to decrease this loss by adjusting the parameters $W_i$.
The \emph{credit assignment problem}~\citep{minskyStepsArtificialIntelligence1961a} states how much of the loss can be attributed to each parameter and how they should be changed accordingly.
A common approach is updating each parameter proportional to the gradient $\delta W_i$ where we use $\delta a=\frac{\partial L(y, y^*)}{\partial a}$ to denote the gradient of the loss $L(y, y^*)$ with respect to a variable $a$.
%
Backpropagation computes this gradient by applying the chain rule in a backward pass over the network.
The gradient of the activations $\delta h_i=W_{i+1}^\top\delta z_{i+1}$ depends on the downstream layer $i+1$ and is, in turn, used to compute the gradient of the weights $\delta W_i = \left(\delta h_i \odot f'(z_i)\right) h_{i-1}^\top$.
The dependency of $\delta h_i$ on the transpose of the forward weights $W_{j}^\top$ and the downstream layers $j>i$ is the root cause of the weight transport and update locking problems which make backpropagation not biologically plausible.

\subsection{Delayed Error Information as Feedback Signal}
\label{sub:delayed_error_info_as_feedback_signals}
We propose an alternative approach to train neural networks by approximating $\delta h_i$ while remaining weight-transport- and update-locking-free.
\emph{\fancynamehighlighted{} (\fancyabrv{})} approximates the gradient $\delta h_i$ using fixed feedback weights $B_i$ and delayed error information $e^{t-1}$ as
\begin{equation}
\label{eq:ffo}
	\delta^\text{\fancyabrv{}} h_i =
	\begin{cases}
		B_i^\top e^{t-1} &\text{for } i \in[1, K-1]\\
		\delta y &\text{for } i=K.
	\end{cases}
\end{equation}
%
We utilize fixed random feedback paths~\citep{lillicrap_random_2016,nokland_direct_2016,frenkel_learning_2021} and replace the layer-wise backpropagation of gradients with a fixed feedback weight matrix $B_i$ to directly pass the error signal to layer $i$.
These weights are randomly initialized and do not change during training.
The forward weights $W_i$ are thus no longer required to approximate the gradient $\delta^\text{\fancyabrv{}} h_i$, making \fancyabrv{} weight-transport-free.

In contrast to prior work, we use delayed error information $e^{t-1}$ from the previous epoch $t-1$ as the error signal.
Computing $\delta^\text{\fancyabrv{}} h_i$ is thus independent of downstream layers and does not need to wait for the forward pass to finish.
This eliminates update locking and allows us to process a sample using only a single forward pass per weight update.
%
To obtain the delayed error, we store the error information from epoch $t-1$ to be used in epoch $t$.
The feedback is thus always delayed by one epoch.
In this paper, we always use sample-wise error information, that is, $e^{t-1}$ contains the delayed error $e^{t-1}[x]$ for each training sample $x$.
We initialize $e^{t-1}$ using the targets $y^*$.
Different types of error information can be used, for example, the loss $\delta y$ (\fancyabrv{}-Loss) or the error $y^*-y^{t-1}$ (\fancyabrv{}-Error).
Additional transformations can be applied to the chosen error information.
For classification tasks, we consider two additional variations: using only the feedback signal of the target class and masking all other entries (\fancyabrv{}-OneHot) or applying an additional softmax to the output $y^{t-1}$ before computing the error information (\fancyabrv{}-Softmax).
Both can be combined with \fancyabrv{}-Loss or \fancyabrv{}-Error.

\begin{algorithm}[t]
    \caption{Training the network described in \ref{sub:backprop} with \fancyabrv{} and SGD.}
    \label{pseudocode}
	\setstretch{1.2}
	\KwIn{Training data $X, Y^*$, maximum epochs $t_{\max}$, and learning rate $\eta$}
	$e^0 \gets Y^*$\FWCommentR{Initialize delayed error with targets}
	\For(\FWCommentF{Iterative training for $t_{\max}$ epochs $t$}){$t\gets1$ {\bfseries to} $t_{\max}$}{
		\ForEach(\FWCommentF{For each training sample $h_0, y^*$}){$h_0, y^*$ \textbf{\upshape{in}} $X$, $Y^*$}{
			\For(\FWCommentF{Forward pass through layers $i$}){$i\gets1$ {\bfseries to} $K$}{
				$h_i \gets f(W_i h_{i-1})$\FWCommentR{Forward step}
				$\delta h_i \gets\delta^\text{\fancyabrv{}} h_i$\FWCommentR{Approximate $\delta h_i$ using \cref{eq:ffo}}
				$\delta W_i \gets \left(\delta h_i \odot f'(z_i)\right) h_{i-1}^\top$\FWCommentR{Compute $\delta W_i$ using the chain rule}
				$W_i \gets W_i - \eta \delta W_i$\FWCommentR{Update weights $W_i$ with SGD}
				}
			$e^t[h_0] \gets y^* - h_K$\FWCommentR{Update delayed error for sample $h_0$}
		}
	}
\end{algorithm}

Having approximated the gradient of the activation $h_i$, the remaining gradients within a layer depend only on layer-local information and are computed with the chain rule using $\delta^\text{\fancyabrv{}} h_i$ instead of $\delta h_i$.
\Cref{pseudocode} gives an example of how \fancyabrv{} can be used to train the fully-connected neural network described in \Cref{sub:backprop} with stochastic gradient descent (SGD).
In each epoch, a sample requires only a single forward pass to approximate the gradients and update the weights in contrast to the forward and backward pass required by backpropagation.
Since \fancyabrv{} only changes how the gradients are computed but does not prescribe a specific update rule, it can be used with various gradient-based optimizers and is not limited to SGD.

\subsection[Theoretical Analysis of \fancyabrvnosuperscript{}]{Theoretical Analysis of \fancyabrv{}}
\label{sub:theoretical_analysis}

To show that \fancyabrv{} decreases the loss, we analyze \fancyabrv{} under the theoretical framework introduced in \cite{nokland_direct_2016}.
This extends upon the feedback alignment principle~\cite{lillicrap_random_2016}, which states that the forward weights $W_i$ align with the fixed feedback weights $B_i^\top$ to allow successful updates via these fixed feedback paths.
Specifically, \citet{nokland_direct_2016} proves the following theorem, which can be applied successively to all layer pairs to demonstrate error-driven training of an entire network.
\begin{theorem}\label{proof:dfa_theorem}
	Given two subsequent hidden layers $i$ and $i+1$ in a feed-forward neural network.
	Let $c_i$ be the backpropagated gradients, $\delta h_j, j\in\{i,i+1\}$ be non-zero update directions prescribed by the feedback paths, and $\frac{\delta h_j}{\|\delta h_j\|}$ be constant for each data point.
	If $L_i=\frac{\delta h_i^\top c_i}{\|\delta h_i\|}>0$, then $-\delta h_i$ is a descending direction to minimize $K_{i+1}=\frac{\delta h_{i+1}^\top h_{i+1}}{\|\delta h_{i+1}\|}$.
\end{theorem}
\citet{frenkel_learning_2021} have shown that $L_i>0$ holds for DRTP under the following assumptions:
A neural network consisting of linear hidden layers $i\in[1,K-1]$ with outputs $h_i=z_i=W_ih_{i-1}$ and a non-linear output layer $K$ with $h_K=\sigma(z_K)$ and $z_K=W_Kh_{K-1}$ using a sigmoid or softmax activation $\sigma$.
The forward-weight matrices $W_i$ are zero-initialized, and $\prod_{i=K}^{k+1}W_i^t$
is right invertible for each step $t>K$.
The random feedback weight matrices $B_i^\top$ have full rank.
The network is trained on a single sample $(x,y^*)$ of a one-hot encoded classification task.
In each update step $t$, the weights are updated according to gradient descent, i.e., $W_i^{t+1}\leftarrow W_i^t-\eta\delta W_i$
with
$\delta W_i=\delta h_ih_{i-1}^\top$ for $i=[1,K-1]$, $\delta W_K=-\frac{1}{C}e^t h_{i-1}^\top$,
and $\eta>0$.
We show that under the same assumptions, $L_i>0$ also holds for \fancyabrv{} after step $t\geq K$ by adapting the alignment proof from \cite{frenkel_learning_2021} to $\delta^\text{\fancyabrv{}} h_i=B_i^\top e^{t-1}$ with $e^{t-1}=y^*-y^{t-1}$.
\begin{theorem}\label{proof:drtp_lemma}
	There are strictly positive scalars $s_{h_i}^t$ and $s_{W_i}^t$ for all hidden layers $i=[1, K-1]$ and a vector $s_{W_K}^t\neq0$ at every update step $t\geq K$ such that
    \begin{minipage}[c]{0.49\textwidth}
        \begin{align}
            W_1^t &= -s_{W_1}^t \delta^\text{\fancyabrv{}} h_1 x^\top\\
            W_K^t &= -s_{W_K}^t \delta^\text{\fancyabrv{}} h_{K-1}^\top\label{eq:lemma_wK}
        \end{align}
    \end{minipage}
    \begin{minipage}[c]{0.49\textwidth}
        \begin{align}
            W_i^t &= s_{W_i}^t \delta^\text{\fancyabrv{}} h_i \delta^\text{\fancyabrv{}} h_{i-1}^\top\\
            h_i^t &= -s_{h_i}^t \delta^\text{\fancyabrv{}} h_i
        \end{align}
    \end{minipage}
\end{theorem}
To prove \Cref{proof:drtp_lemma}, we introduce \Crefrange{proof:drtp_sublemma_non_negative}{proof:drtp_sublemma_syi_positive}.
\begin{lemma}\label{proof:drtp_sublemma_non_negative}
	The scalars $s_{h_i}^t$ and $s_{W_i}^t$ for all hidden layers $i=[1, K-1]$ are non-negative at every update step $t$.
\end{lemma}
\Cref{proof:drtp_sublemma_non_negative} can be proven by induction over $t$ analogously to \cite{frenkel_learning_2021}.
\begin{lemma}\label{proof:drtp_sublemma_swi_positive}
	If the scalars $s_{h_{i-1}}^{t-1}$ are strictly positive for all hidden layers $i=[1, K-1]$ and every update step $t\geq i$, then the scalars $s_{W_i}^t$ are also strictly positive for $t\geq i$ and the vector $s_{W_K}^t$ is non-zero and has the same sign as the error $e^t$ for $t\geq K$.
\end{lemma}
\begin{proof}
	Per \Cref{proof:drtp_sublemma_non_negative}, $s_{W_i}^t\geq 0$ and $s_{h_{K-1}}^{t-1}\geq0$ for all $t$ and per assumption $\eta>0$, $C>0$, and $s_{h_{i-1}}^{t-1}>0$ for $t\geq i$. It follows that $s_{W_1}^t=s_{W_1}^{t-1}+\eta>0$ and $s_{W_i}^t=s_{W_i}^{t-1}+\eta s_{h_{i-1}}^{t-1}>0$ for $i\in\{2,\dots,K-1\}$ and $t\geq i$.
    For the output layer $i=K$, 
    \begin{align}
        W_K^t
        =W_K^{t-1} + \frac{\eta}{C}e^t h_{K-1}^{\top}
        =-s_{W_K}^{t-1} \delta^\text{\fancyabrv{}} h_{K-1}^\top - \frac{\eta}{C}e^ts_{h_{K-1}}^{t-1} \delta^\text{\fancyabrv{}} h_{K-1}^\top
    \end{align}
    and thus $s_{W_K}^t=s_{W_K}^{t-1}+\frac{\eta s_{h_{K-1}}^{t-1}}{C}e^t$.
	With $\sigma$ limiting $y^{t}$ to $(0,1)$ and $y^*$ being one-hot encoded, the delayed error $e^{t-1}$ is non-zero and the sign is constant at all time steps $t$ for samples of the same class.
    The update direction for $s_{W_K}^t$ thus remains the same and by choosing $s_{W_K}^0$ as the zero-vector, it follows that $s_{W_K}^{t}\neq0$ and $\operatorname{sgn}\left(s_{W_K}^{t}\right)=\operatorname{sgn}\left(e^t\right)$ for $t\geq K$.
\end{proof}
\begin{lemma}\label{proof:drtp_sublemma_syi_positive}
	The scalars $s_{h_i}^t$ are strictly positive for all hidden layers $i=[1, K-1]$ at every update step $t\geq i$.
\end{lemma}
\begin{proof}
    Proven by induction over $i$. 
    \begin{align}
        &\text{For } i=1:\qquad s_{h_1}^{t} = s_{h_1}^{t-1}+\eta\|x\|^2>0\\
        &\text{For } i>1:\qquad s_{h_i}^{t} = s_{W_i}^{t}s_{h_{i-1}}^{t} \cdot\|B_{i-1}^\top e^{t-1}\|^2
    \end{align}
    Per induction, $s_{h_{i-1}}^{t-1}$ and $s_{h_{i-1}}^{t}$ are strictly positive.
    Since $B_{i-1}$ has full rank and $e^{t-1}\neq 0$ as shown in \Cref{proof:drtp_sublemma_swi_positive}, $B_{i-1}^\top e^{t-1}\neq 0$ and $\|B_{i-1}^\top e^{t-1}\|^2>0$.
    Thus, the condition for \Cref{proof:drtp_sublemma_swi_positive} is fulfilled and $s_{W_i}^{t}>0$ and therefore $s_{h_i}^{t}>0$.
\end{proof}

\begin{prooftothm2}
	\Cref{proof:drtp_sublemma_non_negative,proof:drtp_sublemma_swi_positive,proof:drtp_sublemma_syi_positive} show the existence of non-negative scalars $s_{h_i}^t$ and $s_{W_i}^t$ that are strictly positive for $t\geq i$.
    They furthermore show the existence of a vector $s_{W_K}^t$ fulfilling \Cref{eq:lemma_wK}
    and prove it is non-zero with $\operatorname{sgn}\left(s_{W_K}^{t}\right)=\operatorname{sgn}\left(e^t\right)$ for $t\geq K$.
\end{prooftothm2}
\begin{theorem}\label{proof:drtp_theorem}
	For every hidden layer $i=[1,K-1]$ and $t\geq K$, $\delta^\text{\fancyabrv{}} h_i$ is a negative scalar multiple of the Moore-Penrose pseudo-inverse of the product of forward matrices and the current error: 
	$\delta^\text{\fancyabrv{}} h_i=-\frac{1}{s_i^t}\left[\prod_{j=K}^{i+1}W_j^t\right]^+e^t$ with $s_i^t>0$.
\end{theorem}
\begin{proof}
    Using \Cref{proof:drtp_lemma}, it follows that
    \begin{align}
		\delta^\text{\fancyabrv{}} h_i &=B_i^\top e^{t-1} = -\frac{1}{s_i^t} \left[\prod_{j=K}^{i+1}W_j^t\right]^+ e^t
    \end{align}
    holds for
    \begin{equation}
        s_k^t=\frac{s_{W_K}^{t^\top} e^t}
        {\left(\prod_{i=k+1}^{K-1}s_{W_i}^t\right) \left(\prod_{i=k}^{K-1} \|B_i^\top e^{t-1}\|^2 \right) \|s_{W_K}^t\|^2}.
    \end{equation}
    Per \Cref{proof:drtp_sublemma_swi_positive,proof:drtp_sublemma_syi_positive}, for every time step $t\geq K$, $s_{W_j}^t>0$ for all hidden layers $j=[1, K-1]$, $s_{W_K}^t\neq0$ and $\operatorname{sgn}\left(s_{W_K}^t\right)=\operatorname{sgn}\left(e^t\right)$.
	Since the $B_j^\top$ have full rank and the delayed error $e^{t-1}$ is non-zero, $B_j^\top e^{t-1}$ is also non-zero, and therefore $\|B_j^\top e^{t-1}\|^2>0$.
    Finally, since $\operatorname{sgn}\left(s_{W_K}^t\right)=\operatorname{sgn}\left(e^t\right)$ and both are non-zero, it follows that $s_{W_K}^{tT}e^t>0$.
    Therefore, all components in $s_i^t$ are positive, and thus $s_i^t>0$ for $t\geq K$.
\end{proof}

This allows us to show the alignment of \fancyabrv{} with backpropagation.
\begin{theorem}
	For every hidden layer $i=[1,K-1]$ and $t\geq K$: $L_i>0$.
\end{theorem}
\begin{proof}
    \begin{align*}
        \left(\delta^\text{BP} h_{i}\right)^{\top} \delta^\text{\fancyabrv{}} h_{i} 
        &=\left(-\frac{1}{C}\left(\prod_{j=i+1}^{K} W_{j}^{\top}\right) e^{t}\right)^{\top} B_{i}^{\top} e^{t-1}\\
        &=-\frac{1}{C}e{^{t}}^{\top}\prod_{j=K}^{i+1} W_{j} \left(-\frac{1}{s_i^t} \left[\prod_{j=K}^{i+1}W_j^t\right]^+ e^t\right)
		=\frac{\|e^t\|^2}{s_i^tC}
    \end{align*}
    As before, $e^t\neq0$, and thus $\|e^t\|^2>0$.
    Furthermore, $C>0$ and per \Cref{proof:drtp_theorem} $s_i^t>0$ for $t\geq K$.
    Therefore, $\frac{\|e^t\|^2}{s_i^tC}>0$
    for all $i=[1,K-1]$, i.e., the updates prescribed by \fancyabrv{} are within 90° of backpropagation, and \Cref{proof:dfa_theorem} applies.
\end{proof}

\subsection{Computation and Memory Demands}
\label{sub:computational}

By eliminating layer-wise error propagation, \fancyabrv{} requires fewer operations per epoch than backpropagation, depending on the input data and selected network architecture.
Consider a fully-connected neural network with $K$ linear layers $i$ computing $h_i=W_ih_{i-1}$ with fixed hidden layer width $n_i=w$ for $i<K$ and output size $n_K=C$.
Thus, $h_i$ has dimension $n_i$, $W_i$ dimension $n_i\times n_{i-1}$, and $B_i$ dimension $n_i\times C$.
Assuming the use of basic multiplication algorithms, both the outer product $xy^\top$ and the matrix-vector product $Ax$ require $ab$~fused multiply-add (FMA) operations with vectors $x\in\mathbb{R}^b, y\in\mathbb{R}^a$ and matrix $A\in\mathbb{R}^{a\times b}$.
Under these assumptions, backpropagation requires $\left(K-2\right)w^2 + Cw$ FMA to compute the gradients $\delta h_i$, while \fancyabrv{} requires $C(K-1)w$ FMA.
For sufficiently large $w$, \fancyabrv{} approaches a near 100\% reduction for computing the $\delta h_i$, resulting in a 25\% theoretical maximum reduction in time across the whole training step, i.e., forward, backward, and update step.
In experimental evaluations, we find that the practically achievable reduction in time per epoch is closer to 15\%.

Backpropagation needs to store the intermediate values $h_i$ from the forward pass until the backward step of layer $i+1$ or perform expensive recomputations from checkpoints.
With a batch size of $b$, this peaks at $\sum_{i=1}^{K-1}n_ib$ elements.
In contrast, \fancyabrv{} does not need to remember intermediate results $h_i$ as gradients are computed immediately during the forward step.
It does, however, come with the additional cost of storing the feedback weights $B_i$ and the delayed error information $e^{t-1}$, consisting of $\sum_{i=1}^{K-1}n_iC$ and $C|X|$ elements, respectively, where $|X|$ is the total number of samples in the training set.
Assuming all values are stored using the same size, e.g. 32 bit, \fancyabrv{} saves $C|X| - (b-C) \sum_{i=1}^{K-1}n_i$
elements.
Overall, this tends to be a slight reduction in memory despite storing additional information as the cached activations outweigh the feedback weights and delayed error.

\section{Evaluation} 
\label{sec:evaluation}

\subsection{Methods} 
\label{sub:methods}

We evaluate \fancyabrv{} on the image classification dataset MNIST~\citep{lecun_gradient_1998} and the regression datasets SGEMM~\citep{nugtren_sgemm_2015,ballester_sgemm_2017} and Wine Quality~\citep{cortezModelingWinePreferences2009}.
All datasets have been standardized.
We train fully-connected neural networks with a hidden layer width of \num{500} neurons and $\tanh$ activation.
For classification, we use one-hot encoding with binary cross-entropy (BCE) loss and a sigmoid activation for the output layer, while mean squared error (MSE) loss is used for regression tasks.
We use Adam~\citep{kingma_adam_2014} with a fixed learning rate determined via hyperparameter optimization; the exact values are provided together with our code linked below.
We compare \fancyabrv{} to multiple other training algorithms: backpropagation as a baseline for traditional training and DFA~\citep{nokland_direct_2016}, DRTP~\citep{frenkel_learning_2021}, and PEPITA~\citep{dellaferrera_error-driven_2022} as bio-inspired alternatives.
We additionally include last-layer-only (LLO), training only the last layer while keeping the hidden layers constant.
All experiments used Python 3.8.6 with PyTorch~\citep{paszke_torch_2018} 1.12.1 and CUDA 11.6.
Our source code is publicly available\footnote{Our source code is available at \href{https://github.com/Helmholtz-AI-Energy/f3}{https://github.com/Helmholtz-AI-Energy/f3}}. For PEPITA, we use the code\footnote{Code for PEPITA available at \href{https://github.com/GiorgiaD/PEPITA}{https://github.com/GiorgiaD/PEPITA}} provided by~\citep{dellaferrera_error-driven_2022}.


\subsection{Comparison to Other Training Algorithms}
\label{sub:evaluation_datasets}

\begin{figure*}[t]
    \begin{center}
    \begin{subfigure}[b]{\textwidth}
        \centering
        \includegraphics[width=\textwidth]{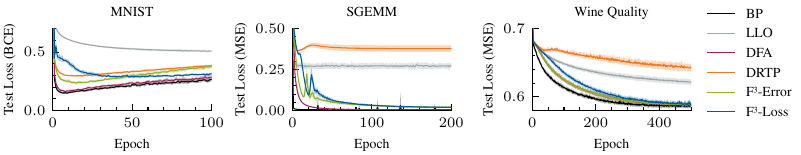}
        \caption{Test loss over time.}
        \label{fig:datasets_results_loss_curves}
    \end{subfigure}
    \begin{subfigure}[b]{\textwidth}
        \centering
        \includegraphics[width=\textwidth]{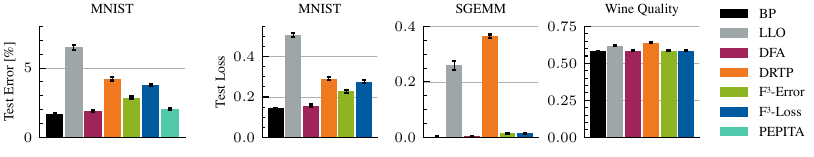}
        \caption{Best top-1-error and test loss.}
        \label{fig:datasets_results_barchart}
    \end{subfigure}
    \caption{
        Training a fully-connected neural network with one hidden layer using different training approaches.}
    \label{fig:datasets_results_overall}
    \end{center}
    \vspace{-0.5cm}
\end{figure*}

\Cref{fig:datasets_results_overall} illustrates the behavior of different training algorithms for multiple datasets and tasks.
Since PEPITA does not use an epoch-wise loss in the same way as the other training approaches, we only report the top-1-error in \Cref{fig:datasets_results_barchart}.
As expected, backpropagation achieves the best performance across all datasets.
Of the biologically inspired training approaches, DFA and PEPITA yield the best performance;
however, both of them are update-locked as they rely on passing the current error backward, either directly or via the second forward pass.
Compared to the similarly plausible DRTP, \fancyabrv{} reduces the gap to backpropagation by a significant fraction.
On MNIST, \fancyabrv{}-Error noticeably outperforms DRTP, reducing the top-1 test error from 4.3\% to 2.7\%.
The regression datasets further emphasize the importance of including magnitude and direction in the feedback signals, as \fancyabrv{} outperforms DRTP by a wide margin, reducing test loss from 0.364 to 0.014 for SGEMM and 0.641 to 0.585 for Wine Quality, while backpropagation achieves 0.003 and 0.583 respectively.
\fancyabrv{} thus reduces the gap to backpropagation by 96.9\% and 96.3\%.
\fancyabrv{} performs significantly better than LLO on all datasets, demonstrating the effectiveness of the biologically plausible update rule for hidden layers.
In summary, \fancyabrv{} notably improves upon the similarly plausible DRTP for both classification and regression tasks, significantly reducing the gap to backpropagation.

\subsection{Impact of the Delayed Error Information and Feedback Weight Initialization}
\label{sub:ablation_study}

\begin{figure*}[tb]
    \begin{center}
    \begin{subfigure}[t]{.575\textwidth}
    \includegraphics[scale=1]{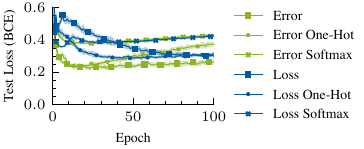}
    \caption{Test loss for \fancyabrv{} using different types of\\delayed error information.}
    \label{fig:error_info}
    \end{subfigure}\hfill
    \begin{subfigure}[t]{.425\textwidth}
        \includegraphics[scale=1]{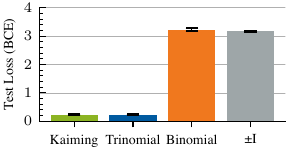}
        \caption{Best test loss using different feedback weight initialization methods.}
        \label{fig:initializations_results}
    \end{subfigure}
    \caption{Comparison of different delayed error information and feedback weight initializations for \fancyabrv{} on MNIST.}
    \end{center}
    \vspace{-0.5cm}
\end{figure*}

\Cref{fig:datasets_results_overall} includes two variants of \fancyabrv{} which use different delayed error information.
For classification, \fancyabrv{}-Error tends to outperform \fancyabrv{}-Loss.
When using MSE loss, the error is equivalent to the loss gradient except for a scalar factor.
In this case, the only difference between \fancyabrv{}-Loss and \fancyabrv{}-Error is the step size; this effect is especially apparent in the loss curve of SGEMM in \Cref{fig:datasets_results_loss_curves}.
In \Cref{fig:error_info} we compare additional transformations of the error information indicated by different markers.
We find that the raw error information and the one-hot versions yield similar results, but one-hot transformation can help smooth the training.
Applying an additional softmax tends to decrease the resulting quality as measured by an increased test loss.

We further explore different approaches to initializing the feedback weights $B_i$ and their impact on the training.
We compare a Kaiming-He uniform distribution~\citep{he_delving_2015} with two discrete uniform distributions with values $\{-1, 0, 1\}$ (Trinomial) and $\{0, 1\}$ (Binomial) and repeating the identity matrix $I$ along the larger dimension with alternating signs ($\pm I$).
\Cref{fig:initializations_results} shows the test loss for the different initialization methods.
We observe the trinomial initialization to perform as well as Kaiming uniform while both binomial and $\pm I$ initialization significantly increase the loss.
These results illustrate that algorithms based on random feedback paths do not depend on a specific initialization of the feedback weights, but certain characteristics, such as multiple non-zero values and different signs in the columns of the two-dimensional feedback weight matrix, are necessary to promote effective training.

\subsection{Performance on Different Model Architectures}
\label{sub:eval_different_models}
\label{sub:scaling_with_depth}

\begin{figure*}[t]
    \begin{center}
    \begin{subfigure}[t]{.63\textwidth}
    \includegraphics[scale=1]{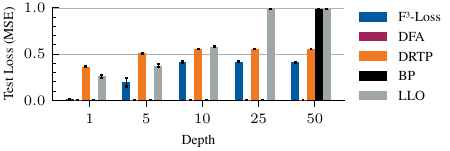}
    \caption{SGEMM for different network depths.}
    \label{fig:depth_results}
    \end{subfigure}\hfill
    \begin{subfigure}[t]{.35\textwidth}
        \includegraphics[scale=1]{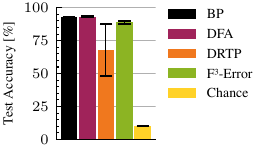}
        \caption{MNIST and ViT.} 
        \label{fig:transformer_barchart}
    \end{subfigure}
    \caption{Performance on different model architectures.}
    \label{fig:depth_and_transformer}
    \end{center}
    \vspace{-0.5cm}
\end{figure*}

\Cref{fig:depth_results} shows the scaling behavior with increasing network depth.
We find that the direct feedback pathways~\citep{nokland_direct_2016} help alleviate problems typically arising with very deep networks~\citep{srivastava_advances_2015}, even when using activations prone to vanishing gradients.
As expected from \citet{nokland_direct_2016}'s results, DFA is entirely unaffected by the network depth, while for DRTP and \fancyabrv{}, performance decreases slightly but much slower than with backpropagation.
This makes the biologically plausible algorithms DFA, DRTP, and \fancyabrv{} more robust and able to train even deep, 100-layer networks, while backpropagation fails to train networks with more than 25 layers without additional remedies like specific activation functions or initializations.

Biologically plausible approaches have typically focused on simpler architectures like fully-connected and convolutional networks.
To explore the applicability of such approaches to more complex architectures, we extend \fancyabrv{}, and with it, DFA and DRTP, to Transformer networks~\citep{vaswaniAttentionAllYou} using a simplified Vision Transformer (ViT)~\citep{dosovitskiy_image_2020} with eight heads and four Transformer blocks with tanh activation.
We use patch embedding but no class token or positional encoding on the input and feed all output tokens into the MLP classifier head.
For the biologically inspired algorithms, we treat each Transformer block as one layer which receives an approximate gradient for its output and computes internal gradients independently of other layers.
\Cref{fig:transformer_barchart} shows the best test accuracy on MNIST achieved throughout \num{100} epochs.
We observe that both DFA and \fancyabrv{} can train this Vision Transformer to a similar accuracy as backpropagation, while DRTP offers a significant improvement over random chance but does not reach the same accuracy as the other algorithms tested.
This demonstrates that biologically inspired algorithms like \fancyabrv{} and DFA can successfully train Transformer networks.
However, their training is sensitive to the input and output encoding and the type of activations used.
Reaching state-of-the-art performance will thus require further fine-tuning.

\section{Conclusion} 
\label{cha:conclusion}

Backpropagation relies on concepts infeasible in natural learning processes, rendering it an effective yet biologically implausible and computationally expensive approach to training neural networks.
In this paper, we introduce \fancynamehighlighted{} (\fancyabrv), a novel algorithm for backpropagation-free training of multi-layer neural networks using a feed-forward approximation of intermediate gradients based on delayed error information.
\fancyabrv{} solves both the weight transport and update locking problems and significantly improves the predictive performance compared to previous update-locking-free algorithms, reducing the gap to backpropagation by up to 56\% for classification and 96\% for regression tasks.
This paper successfully demonstrates the application of \fancyabrv{} to fully-connected and Transformer networks. 
There are multiple avenues for future research, such as further fine-tuning to scale up to large-scale tasks or extending \fancyabrv{} to more model architectures.
In theory, \fancyabrv{} can be applied to convolutional networks, but further research is necessary to preserve the advantages of weight sharing for the feedback weights and make them independent of the input size.
While some implausibilities remain, such as the spiking problem, Dale’s Law, and supervised learning in general~\citep{bengio_towards_2016,ecclesElectricalChemicalTransmission1976}, \fancyabrv{} is an important step towards more biologically plausible learning algorithms and can even improve computational efficiency.
%
By releasing update locking and thus the inter-layer dependencies when computing the gradients, \fancyabrv{} can update the network's parameters during the forward pass.
As a result, \fancyabrv{} eliminates the need for the computationally expensive backward pass, thus requiring fewer operations and removing the need to buffer or recompute activations.
Beyond that, it opens up promising possibilities for parallel training setups like pipeline parallelism by reducing the amount of communication and synchronization between different layers.
Furthermore, \fancyabrv{} enables on-device training using highly promising neuromorphic devices, which would simplify the training process significantly and has the potential to economize compute resources and energy.


\begin{credits}
\subsubsection{\ackname} This work is supported by the Helmholtz Association Initiative and Networking Fund under the Helmholtz AI platform and the HAICORE@KIT grant.

\subsubsection{\discintname}
The authors have no competing interests to declare that are
relevant to the content of this article.
\end{credits}


\bibliography{references}

\end{document}